\documentclass[10pt,twocolumn,letterpaper]{article}

\usepackage[pagenumbers]{cvpr} 

\usepackage{svg}
\usepackage{siunitx}
\usepackage{pythonhighlight}
\usepackage{tabularx}

\usepackage{array}
\newcolumntype{L}{>{\raggedright\arraybackslash} p{0.20\textwidth}}
\newcolumntype{C}{>{\centering\arraybackslash} p{0.07\textwidth}}
\newcolumntype{D}{>{\centering\arraybackslash} p{0.19\textwidth}}

\usepackage[accsupp]{axessibility}

\definecolor{cvprblue}{rgb}{0.21,0.49,0.74}
\usepackage[pagebackref,breaklinks,colorlinks,allcolors=cvprblue,breaklinks=true,,bookmarks=false]{hyperref}

\def\paperID{4949} 
\def\confName{CVPR}
\def\confYear{2025}

\title{MozzaVID: Mozzarella Volumetric Image Dataset}

\author{Pawel Tomasz Pieta
\and
Peter Winkel Rasmussen
\and
Anders Bjorholm Dahl
\and
Jeppe Revall Frisvad
\and
Siavash Bigdeli
\and
Carsten Gundlach
\and
Anders Nymark Christensen \\ 
{\tt\small \{papi, pwra, abda, jerf, sarbi, cagu, anym\}@dtu.dk}\\ 
Technical University of Denmark, 
Kgs. Lyngby, Denmark\\
}

\begin{document}
\maketitle
\begin{abstract}
Influenced by the complexity of volumetric imaging, there is a shortage of established datasets useful for benchmarking volumetric deep-learning models. As a consequence, new and existing models are not easily comparable, limiting the development of architectures optimized specifically for volumetric data. To counteract this trend, we introduce MozzaVID -- a large, clean, and versatile volumetric classification dataset. Our dataset contains X-ray computed tomography (CT) images of mozzarella microstructure and enables the classification of 25 cheese types and 149 cheese samples. We provide data in three different resolutions, resulting in three dataset instances containing from 591 to 37,824 images. While targeted for developing general-purpose volumetric algorithms, the dataset also facilitates investigating the properties of mozzarella microstructure. The complex and disordered nature of food structures brings a unique challenge, where a choice of appropriate imaging method, scale, and sample size is not trivial. With this dataset, we aim to address these complexities, contributing to more robust structural analysis models and a deeper understanding of food structure. The dataset can be explored through: \url{https://papieta.github.io/MozzaVID/}
\end{abstract}    
\section{Introduction}
\label{sec:intro}

Volumetric images reveal the internal shape and structure of objects, making them important for a wide range of fields such as medical imaging~\cite{singh2020a, doi2007a, heimann2009a}, material science~\cite{salvo2010a, maire2014a,plessis2019a}, paleontology~\cite{sutton2008a,cunningham2014a} and food science~\cite{Schoeman2016, Du2019}.
With the growing use of volumetric data, many image analysis algorithms have been developed to work well with this modality. Most notably, the versatility and power of deep learning methods have caused them to dominate research contributions, especially in the medical field~\cite{zhou2021a, litjens2017a, Shen2017, mridha2022a}. Together with this development, an interest in obtaining comprehensive, well-curated volumetric datasets~\cite{walsh2021a, setio2017a, Menze2015, mota2024} has grown proportionally.

\begin{figure}[!t]
\centering
\includegraphics[width=0.98\columnwidth]{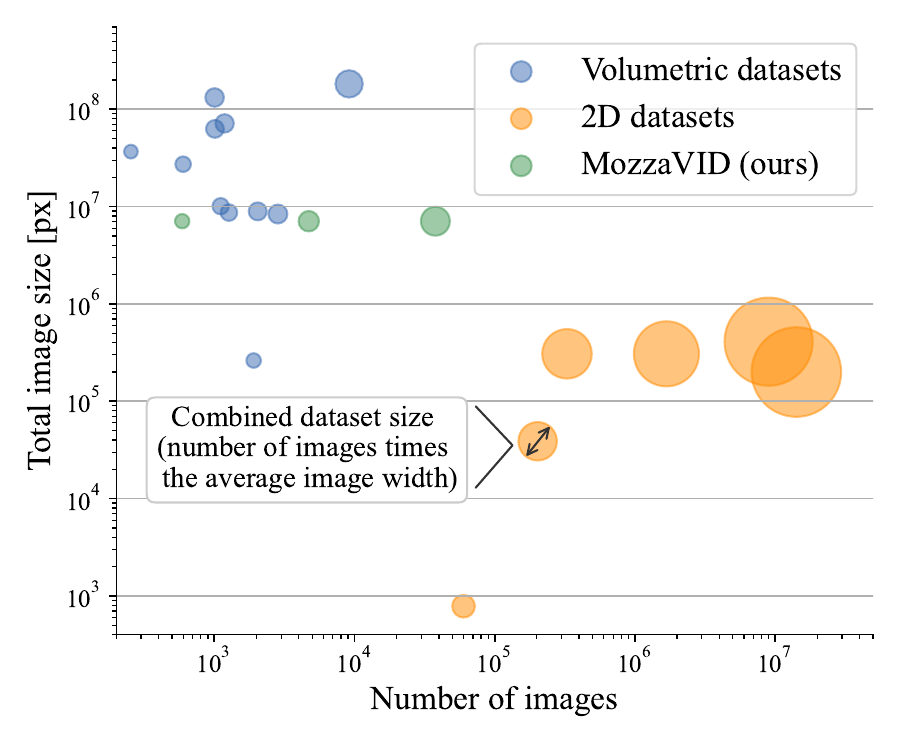}
\caption{
Comparison of typical volumetric and 2D dataset sizes. The three instances of the proposed MozzaVID dataset form a bridge between the two groups while maintaining the volume sizes known from other volumetric datasets. A complete overview of the visualized datasets can be found in \cref{tab:datasets,tab:datasets_2D}.}
\label{fig:2D_3D_dataset_sizes}
\end{figure}

While the availability of volumetric datasets grows, their scope is still far from that known in 2D datasets. Volumetric imaging requires complex acquisition setups, and typically generates few, very large image instances (between 1 and 100). Additionally, its most common area of use --- medical research --- poses further practical constraints on data collection and sharing \cite{national2016principles}. As a result, the largest existing volumetric datasets are limited to a few moderately-sized collections (containing between 100 and \num{10000} volumes)~\cite{BRATS_new,lidc-idri,21PAMI-SANet,heller2020kits19challengedata300}. However, these larger datasets usually target a complex scientific question that is difficult to address within a single deep-learning task. In contrast, even the smallest of widely recognized 2D datasets contain more than \num{60000} images~\cite{Deng2012,Lin2014,Liu2015}, and their primary objective is often to support general-purpose method development, which is reflected in their design.

In 2D model development, it has become a standard practice to evaluate proposed architectures on a set of benchmark datasets~\cite{he2016a,sandler2018a,liu2022a,vit,swin,xie2017a,tan2019_new}, demonstrating their versatility across different tasks. However, the limitations of existing volumetric datasets make it difficult to conduct similar evaluations. As a result, the most impactful contributions to the field tend to focus on adapting State-of-the-Art (SotA) 2D architectures to solve specific problems defined by a single dataset~\cite{i2016a,hatamizadeh2022a,xie2021a,tang2022a,wasserthal2023a}. While this process can be effective at times, the properties of volumetric datasets make it likely that models optimized specifically for this modality may outperform those that were simply adapted from 2D~\cite{Zheng2015,kim2024a, graham2018a, zhou2023a}.

To facilitate the development and benchmarking of volumetric models, we introduce the mozzarella volumetric image dataset (MozzaVID). It consists of 591 synchrotron X-ray computed tomography (CT) scans of mozzarella cheese microstructure, enabling the classification of 25 cheeses and fine-grained classification of 149 samples. Mozzarella has an anisotropic microstructure that is highly disordered~\cite{Bast2015, Feng2023}, allowing for arbitrary splitting of the original scans with a low risk of introducing bias or losing crucial information. This property enables creating datasets with varying sample sizes and resolutions, which we demonstrate by proposing three dataset splits: containing \num{591}, \num{4728}, and \num{37824} samples.

Most specifically, this dataset is targeted towards developing and evaluating methods for deep learning-based analysis of food structures. Mozzarella can serve as a model system for a range of protein-based foods containing dispersed fat and/or water regions, including meat~\cite{bailey1972a}, dairy~\cite{everett2017a}, or their plant-based analogues~\cite{samard2019a,soezeri2025a}. All of those foods feature complex, disordered microstructures that have a direct impact on their functional and sensory properties. 
With 34\% of greenhouse gas emissions linked to food~\cite{crippa2021a}, a detailed understanding of structural characteristics is crucial for developing environmentally friendly alternatives to known structured foods that are also pleasant to eat~\cite{bhat2011a,godoi2016a,dobson2023a}.

More broadly, mozzarella shares clear visual and conceptual similarities with other types of organic and medical volumetric data~\cite{Andersson2020,kim2024b,schaad2017a,christen2015a}, suggesting that model performance on MozzaVID can serve as a proxy for a wide variety of common analysis objectives. With two classification targets and three dataset splits, MozzaVID supports evaluation across different levels of feature detail and varying data availability constraints. This positions MozzaVID as a dataset that bridges the gap between large, method-oriented 2D datasets and typical volumetric datasets, as shown in \cref{fig:2D_3D_dataset_sizes}.

The microstructural variation of the MozzaVID samples is induced by a combination of the chemical composition and the processing parameters of the raw cheese curd. These parameters, along with rheological and functional measurements, form metadata, which will be published in a simplified form with the dataset.

The separation into cheese types and samples forms a hierarchy, where each cheese comprises 6 samples, with 4 scans performed within each sample, where the only difference between those is their spatial position. With this separation, we expect the samples to exhibit three levels of similarity:  superficial similarity between samples produced with comparable recipes,  moderate similarity between samples from the same cheese, and strong similarity between scans of the same sample.

Our experiments confirm the outlined relationships -- with the biggest dataset instance we can obtain close to perfect classification of the 25 cheeses, as well as high accuracy on the 149 samples. By investigating the embeddings of the classifiers, we further show that they learn to arrange the classes into groups of cheese types with similar parameters, and create a latent space that accurately covers the extent of the structural variations of the samples. This shows that through classification alone we can investigate and quantify the variability and relationships of the analyzed structures.

Our contributions can be summarized as follows:
\begin{itemize}
    \item We introduce MozzaVID, a large and versatile dataset that bridges the gap between established 2D benchmark datasets and biggest volumetric datasets,
    \item We compare the performance of SoTA architectures on MozzaVID both in 2D and 3D, emphasizing the need for a specialized volumetric-oriented deep learning research,
    \item We explore the capabilities of MozzaVID in describing and explaining variability in mozzarella microstructure, highlighting its significance for deep learning-based analysis of structured foods.
\end{itemize}
\section{Related work}
\label{sec:related_work}

\begin{table*}[t]
\centering
\caption{Existing biggest and most recognized volumetric datasets.}
\footnotesize
\label{tab:datasets}
\begin{tabularx}{\textwidth}{llllCDC}
\\[-4ex]
\toprule
Dataset      & \# of volumes & Volume size & Primary application                   & Directly accessible & Task & \# of labels                              \\ \midrule
ADAM         & 254           & 512$\times$512$\times$140 (max) & Medical: brain MRI           & No                 & Detection, segmentation & 2 \\
BraTS        & 50--2,040       & 240$\times$240$\times$155 & Medical: brain MRI          & No                 & Segmentation & 4 \\
KiTS         & 599           & 512$\times$512$\times$104 & Medical: kidney CT           & Yes                & Segmentation, classification & 3 \\
LIDC-IDRI    & 1,010          & 512$\times$512$\times$(65--764) & Medical: thoracic CT         & Yes                 & Detection, classification & 2 \\
MosMedData   & 1,110          & 512$\times$512$\times$(36--41) & Medical: chest CT            & Yes                & Segmentation, classification & 5 \\
CTSpine1k    & 1,005          & 512$\times$512$\times$(349--659) & Medical: spine CT            & No                 & Segmentation & 25 \\
CTPelvic1k~\cite{CTPelvic1K_new}   & 1,184          & 512$\times$512$\times$273 (mean) & Medical: pelvic CT           & Yes                & Segmentation & 5 \\
OASIS        & 2,842     & 256$\times$256$\times$128 & Medical: brain MRI           & No                 & Segmentation, classification & 4 \\
PN9          & 8,798          & Not reported & Medical: thoracic CT         & No                 & Detection & 9 \\
ATLAS v2.0~\cite{Liew2022}   & 1,271          & 233$\times$197$\times$189 & Medical: brain MRI           & Yes                &  Segmentation & 2 \\
MedMNIST 3D & 1,633--1,908 & 64$\times$64$\times$64 & Method development & Yes & Classification & 2/11 \\
BugNIST     & 9,154 + 388      & 900$\times$450/650$\times$450/650  & Method development    & Yes                & Detection under domain shift, classification & 12 \\
\midrule
MozzaVID (ours) & \numrange{591}{37824}      & 192$\times$192$\times$192 & Food science                 & Yes                & Classification & 25/149 \\ \bottomrule
\end{tabularx}
\normalsize
\end{table*}
\begin{table*}[t]
\centering
\caption{A subset of the most recognized 2D datasets as well as food-oriented 2D datasets.}
\footnotesize
\label{tab:datasets_2D}
\begin{tabularx}{\textwidth}{lllcl}
\\[-4ex]
\toprule
Dataset & \# of images & Image size & Primary application &  Task\\ \midrule
MNIST, Fashion-MNIST & \num{60000} (each) & 28$\times$28 & Method testing &  Classification\\
CelebA & \num{202599} & 178$\times$218 & Method development & Face attribute recognition, face recognition, face detection \\
COCO & $\sim$\num{328000} & 640$\times$480 & Method development &  Object detection, keypoint detection, panoptic segmentation\\
ImageNet~\cite{Deng2009} (2014) & \num{14197122} & 482$\times$415  & Computer vision & Object detection, image classification\\
Open Images~\cite{Kuznetsova2020} & $\sim$\num{9000000} & Unspecified & Computer vision &  Object and visual relationship detection, instance segmentation\\
MSLS~\cite{Warburg_2020_CVPR} & $\sim$\num{1680000} & 480$\times$640 & Computer vision & Visual place recognition\\
FoodSeg103/152 & \num{7118}/\num{9490} & Unspecified & Computer vision & Segmentation\\
Recipe1M+ & \num{13000000} & Unspecified & Computer vision & Multimodal learning \\
\bottomrule
\end{tabularx}
\normalsize
\end{table*}

Volumetric datasets vary widely in imaging methods, sample sizes, and application domains. By far, the biggest and most diverse group consists of instances with a small volume count and no or limited annotation available. The goal of these datasets is a general analysis of the imaged object --- either for domain-specific research or for advancing imaging technology. The two most notable mentions in this category are TomoBank~\cite{de2018a} --- a large repository of lab-scale and synchrotron CT datasets, and The Human Atlas project~\cite{deng2024} --- a repository of hierarchical phase-contrast CT of human organs.

Annotated volumetric datasets, particularly those targeted for deep learning applications, are predominantly found in the medical field. In this field, bigger datasets are available for research in specific diagnostic areas, typically consisting of \numrange{100}{1500} volumes. Frequently, these datasets are linked to a public challenge, demonstrated by well-recognized instances such as BraTS~\cite{BRATS_new,Menze2015,Bakas2017}, KiTS~\cite{heller2020kits19challengedata300, heller2021a}, or LIDC-IDRI/Luna16~\cite{lidc-idri,setio2017a}. Despite being substantial by volumetric imaging standards, they are still relatively small for deep learning purposes. Notably, the PN9 dataset~\cite{21PAMI-SANet} provides a larger sample of \num{8798} lung CT scans. Outside the medical field, dataset availability is much more limited, with BugNIST~\cite{jensen2024a} being a rare example, containing \num{9544} scans of common insects. A full overview of the biggest and most recognized volumetric datasets can be found in \cref{tab:datasets}.

A majority of the existing datasets (especially those within the medical field) focus on segmentation or detection as the primary task. A smaller subset also offers classification targets (KiTS, MosMedData~\cite{morozov2020mosmeddatachestctscans_new}, OASIS~\cite{OASIS}), often as an extension to the main target. MedMNIST 3D~\cite{Yang2023} and BugNIST are examples of large classification-oriented datasets. However, in the case of BugNIST, the baseline classification task is quite trivial, serving as a preliminary step before the main, domain shift task. 

A significant limitation of many of the medical datasets is their accessibility~\cite{national2016principles}. Some require registering an account and/or agreeing to terms and conditions (BraTS, ADAM~\cite{ADAM}, OASIS), while others are only fully accessible after personally contacting the publisher (CTSpine1k~\cite{deng2024}, PN9). 
From a methodological perspective, existing datasets pose further challenges. Many suffer from class imbalance, various sources of annotation and representation bias, as well as high specificity of the investigated problem, making them too constrained to serve as a generic baseline~\cite{Li2023}. Consequently, most volumetric deep learning research relies on unique, often case-specific datasets, limiting model generalizability and comparability across studies.

Serving as a context to presented characteristics, \cref{tab:datasets_2D} lists a subset of the most recognized 2D datasets. In this summary, even the smallest of the datasets (MNIST~\cite{Deng2012}, Fashion MNIST~\cite{Xiao2017}) contain many more data instances than any of the existing volumetric counterparts. It is also worth noting that many of these 2D datasets were made primarily for testing or development of new methods (MNIST, Fashion MNIST, CelebA~\cite{Liu2015}, COCO~\cite{Lin2014}). This approach is further reflected in the design of these datasets, through their simplicity, image size, curated setup, and clear task definition.

The dataset representation within food science is especially limited. 
Existing food-focused datasets, such as FoodSeg103/152~\cite{Wu2021} or  Recipe1M+~\cite{Marin2021}, are restricted to 2D photos of food products or prepared dishes. To date, no large, publicly accessible dataset targeted specifically at food structure, either in 2D or 3D, has been released. The high cost and complexity of performing imaging studies on food microstructure mean that most published work relies on a handful of case-specific images that are neither easily accessible nor practically reproducible.
\section{Data} \label{sec:data}

The MozzaVID dataset consists of high-resolution CT images of 25 mozzarella cheese types with diverse functional properties. It is the first published instance of synchrotron-based CT imaging of mozzarella, providing detailed and high-quality images of its 3D microstructure.

\subsection{Acquisition and preprocessing} \label{sec:data_acq_preproc}

The anisotropic structure of mozzarella is created in the cooking-stretching step, where the cheese curd is heated and simultaneously kneaded with a rotating screw~\cite{feng2021a}. Investigated mozzarella types are made with varying cooking temperatures, speeds of the screw as well as optional additives. This experimental design is made to represent a set of realistic recipes and capture the range of potential structural variability of mozzarella. Importantly, three pairs of cheese were prepared with the same recipe, and the Cagliata cheese (class 25) was produced without stretching the curd, making it fully isotropic. Each cheese type was characterized with rheological and chemical measurements, enabling future structural analysis with detailed metadata (used in simplified form due to confidentiality). The samples were stored frozen and thawed directly before imaging.

Mozzarella cheese is a challenging target for CT imaging because its two primary components -- proteins and fats have a similar, low X-ray attenuation coefficient, which results in prolonged scan times and sample thermal instability. Mozzarella can be scanned using laboratory micro-CT, but the resulting images are limited in number and very noisy~\cite{Feng2023}. A potential solution is to use a synchrotron X-ray light source. Although it is not applicable for routine data collection, it provides X-ray radiation of high flux and coherence, enabling fast scans at high resolution and low noise levels. 

\begin{figure}[!t]
\centering
     \begin{subfigure}[b]{0.42\columnwidth}
         \includegraphics[width=\columnwidth]{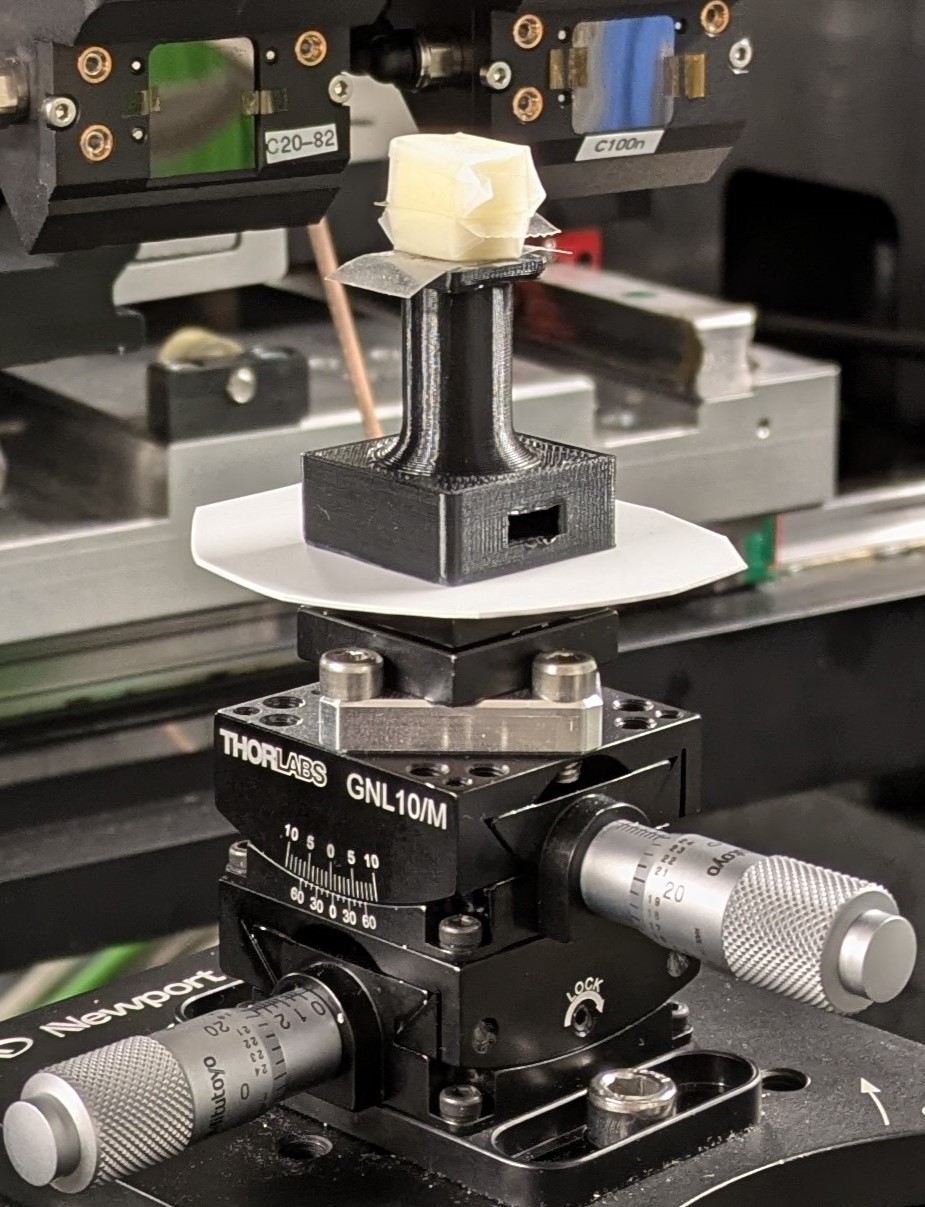}
     \end{subfigure}
     \hspace{0.04\columnwidth}
     \begin{subfigure}[b]{0.51\columnwidth}
        \centering
         \includegraphics[width=\columnwidth]{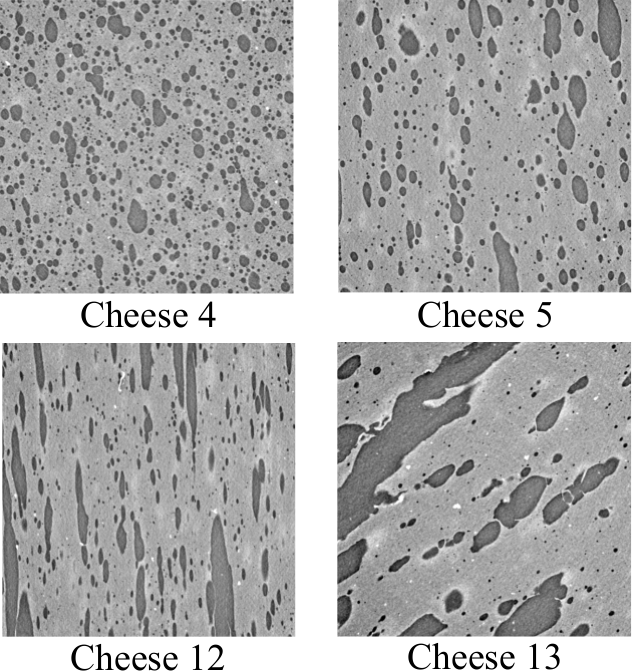}
     \end{subfigure}
\caption{Mozzarella samples wrapped in parafilm and mounted for scanning (left). Structure variability demonstrated by 2D slices from four different cheese types (right). Light areas represent the protein matrix, while the dark areas are the fat globules/domains.}
\label{fig:scaning_and_example_scans}
\end{figure}

The measurements were conducted at the DanMAX beamline of the MAX IV synchrotron. Six samples were prepared from each cheese type by cutting out \SI{1}{\centi\metre} cubes and wrapping them in parafilm, summing up to a total of \num{150} samples. Within each sample, four local tomography scans were taken, resulting in a total of \num{600} individual scans. All the scans were performed using the energy of \SI{20}{\kilo\electronvolt} and exposure time of \SI{1.5}{\milli\second}. \num{2601} unique projections were taken at \SI{0.55}{\micro\metre} pixel size and 2356$\times$2688 pixel resolution. The final scan width was approximately \SI{1.3}{\milli\metre}. The scanning setup is introduced in \cref{fig:scaning_and_example_scans}.

Of the original reconstructed scans, 9 were discarded due to artifacts that heavily compromised image quality. In particular, a whole sample from cheese 4 was discarded, lowering the total number of samples to 149. The final set of 591 scans was cropped to the shape of 1601$\times$1601$\times$2156 (XYZ), and their histograms aligned by segmenting out the fat and protein and standardizing their intensities. This processed data was then treated as a raw input for the dataset.

\subsection{Dataset preparation}

The extracted raw scans (\cref{fig:scaning_and_example_scans}) present a detailed and clear description of the mozzarella structure, but their resolution is too high for most deep-learning algorithms. To address this issue, we propose a pipeline that downsamples the volumes and splits them to create more data instances. As previously mentioned, the internal structure of mozzarella does not contain specific macroscopic shapes or boundaries, allowing the samples to be split into smaller volumes without introducing bias. Additionally, the lack of repeating patterns or monotonous structures further minimizes the risk of ineffective splits. The practical limiting factors are the maximum split number and maximum effective pixel size -- they need to be conservative enough to retain a meaningful representation of the structure. Excessive splitting could result in sub-volumes that are too small to describe relevant features, while excessive downsampling may lead to a loss of critical structural details.

\begin{figure}[!t]
\centering
\includegraphics[width=\columnwidth]{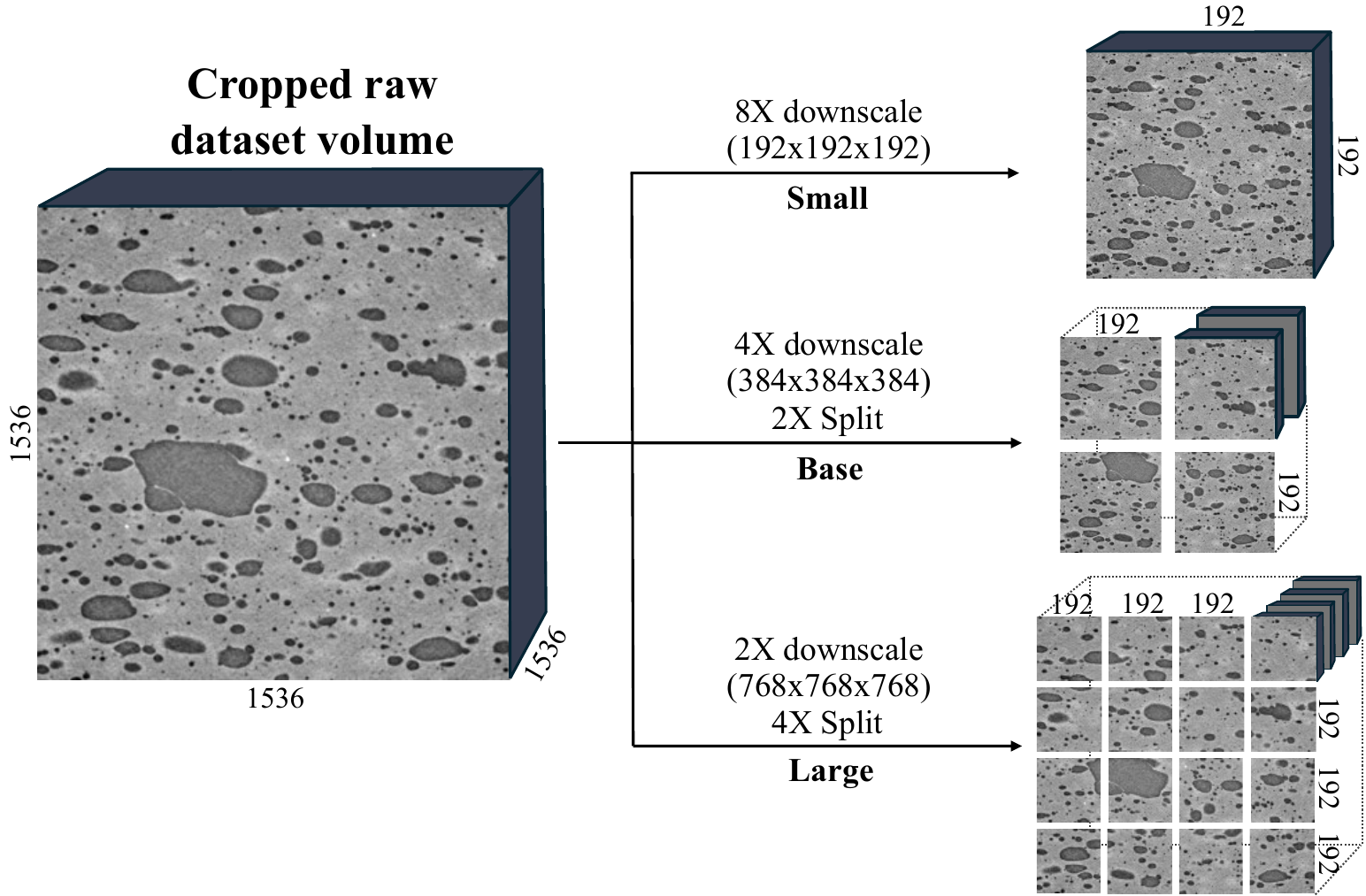}
\caption{
Sketch of the three proposed dataset configurations. The raw volume is downscaled and split, ensuring that in each case, the final volumes have the shape of 192 cubed.}
\label{fig:daatset_instance_creation}
\end{figure}

Three main configurations of the dataset were prepared. Starting by limiting the volume size to a central cube of 1536 pixel width, the instances are then defined as:
\begin{enumerate}
    \item \textbf{8X-1X (Small)} -- each volume downscaled 8-fold, preserving the original count of 591 volumes,
    \item \textbf{4X-2X (Base)} -- each volume downscaled 4-fold and split into equal 8 sub-volumes (2 in each dimension), resulting in 4,728 volumes,
    \item \textbf{2X-4X (Large)} -- each volume downscaled 2-fold and split into 64 sub-volumes (4 in each dimension), resulting in \num{37824} volumes.
\end{enumerate}
The output resolution of all the configurations is standardized at 192$\times$192$\times$192 voxels. \cref{fig:daatset_instance_creation} illustrates the dataset preparation process.

We settle on the outlined approach as it is simple, creates volumes of manageable size, and aligns well with the goals of the dataset. The number of volumes in the Base and Large instances is well suited for the targets of 25 and 149 classes respectively, while also enabling the exploration of the influence of scale on the accuracy. Although the Small instance is challenging for most non-trivial deep learning applications, it reflects the typical size of volumetric datasets, which rarely exceed this number. Consequently, potential volumetric methods have to be ready to address such a limitation.

\section{Experiments} \label{sec:experiments}

Within the proposed dataset, there is a hierarchy of targets and approaches that can be explored, forming three distinct ablation studies: classification granularity, dataset configuration, and dimensionality. These factors combine to form ten experimental setups that we evaluate on a set of benchmark architectures.
To assess the potential of classification in learning the space of possible mozzarella structures, we also investigate the embeddings of trained models and compare their distribution to the properties known from the experimental design of available cheese types.

\subsection{Model selection and training setup}
\label{sec:exp_models}

We selected five widely recognized model architectures within convolutional neural networks and transformers to evaluate performance across the dataset variations.  Due to the volume instance size (which roughly corresponds to a 2K RGB image) and subsequent memory limitations, most models are in their small/medium size version:
\begin{itemize}
    \item \textbf{Convolutional Neural Networks (CNNs)}:
        \begin{itemize}
            \item \textit{ResNet50}~\cite{he2016a} -- a standard benchmark model, sourced from TorchVision~\cite{torchvision2016} and adapted to 3D by replacing 2D components (\eg convolutions, pooling) with their 3D versions.
            \item \textit{MobileNetV2}~\cite{sandler2018a} -- an alternative lightweight model, potentially fitting smaller volumetric datasets better. Chosen in the V2 version due to the availability of its 3D implementation at the 3D-CNN-PyTorch repository~\cite{3d_cnn_pytorch}. A 2D version was sourced from TorchVision.
            \item \textit{ConvNeXt-S }~\cite{liu2022a} -- a recent CNN architecture inspired by transformers. Sourced from the official repository and adapted to 3D in the same manner as ResNet50.
        \end{itemize}
    \item \textbf{Transformers}:
        \begin{itemize}
            \item \textit{ViT-B/16}~\cite{vit} -- a foundational vision transformer. Sourced from the vit-pytorch repository~\cite{vit-pytorch},
            \item \textit{Swin-S}~\cite{swin} -- a hierarchical vision transformer that is more sensitive to multi-scale information. A 2D version was sourced from TorchVision, a 3D version was extracted from the components of SwinUNETR~\cite{hatamizadeh2022a}.
        \end{itemize}
\end{itemize}

All models were trained using the AdamW optimizer~\cite{adamw} and an effective batch size of 32. The learning rate was fine-tuned for all models apart from the biggest 3D instances, where it was set to \num{e-4} (details in the \emph{Supplementary Material}).  Data augmentation was limited to random flipping along the X and Y axes to preserve the spatial structure. All images are normalized using mean and standard deviation extracted from the raw data. The data is split into training, validation, and test sets using separate splitting approaches for the coarse and fine targets (details in the \emph{Supplementary Material}). We use cross-entropy loss for training and test set accuracy as the final performance evaluation metric. For each experimental setup, models were trained with an early stopping criterion based on validation loss. The CNNs were trained with a 30-epoch patience, while the transformer-based models, which generally converge more slowly, used a 50-epoch patience. If a model did not converge within five days, its training was stopped at the currently best result.

\subsection{Ablation studies}

\subsubsection{Granularity and dataset configuration}

We evaluate two classification targets within the dataset: coarse-grain, corresponding to the 25 cheese types, and fine-grain, corresponding to the 149 individual cheese samples. For coarse-grain classification, we use all three dataset configurations: Small, Base, and Large. For fine-grain classification, only the Base and Large configurations are used. In the Small instance, the ratio of sample size to class number is too small.

\subsubsection{Dimensionality}

To effectively investigate the influence of the volumetric representation, we compare models trained on full 3D volumes with those trained on 2D slices, simulating simpler imaging techniques, such as microscopy. In 2D imaging, the most accurate anisotropy measurements are obtained when the imaging plane is aligned parallel to the fiber direction, and in the case of mozzarella, this direction can be inferred with the naked eye. In our dataset, due to the sample preparation, the fiber direction is always roughly perpendicular to the Z-axis. This means that Z-axis slices are an approximate representation of a 2D imaging approach. To provide a better overview of the structure in each volume, and fit to typical 3-channel 2D architectures, we choose three such slices from each volume: at 25\%, 50\%, and 75\% of the height. For a fair comparison, no pre-trained weights are used on either 2D or 3D models.

\begin{table*}[t!]
\caption{Classification accuracy of the trained models. The best results for each granularity are highlighted in bold, while the underline marks the best results in each experimental setup. Models marked with an asterisk did not converge within the time limit.}
\centering
\begin{tabular}{@{}l@{\hskip 20pt}cc|cc|cc@{\hskip 30pt}cc|cc@{}}
\toprule
Granularity   & \multicolumn{6}{c@{\hskip 35pt}}{Coarse}  & \multicolumn{4}{c}{Fine} \\ \midrule
Split         & \multicolumn{2}{c}{Small}  & \multicolumn{2}{c}{Base} & \multicolumn{2}{c@{\hskip 30pt}}{Large} & \multicolumn{2}{c}{Base} & \multicolumn{2}{c}{Large} \\ \midrule
Dimensionality & 2D     & 3D     & 2D     & 3D     & 2D     & 3D     & 2D     & 3D     & 2D     & 3D   \\ \midrule
ResNet50       & 0.381  & \underline{0.763}  & 0.741  & \underline{0.957}  & \underline{0.777}  & \underline{\textbf{0.973}}  & \underline{0.563}  & 0.686  & 0.770  & \underline{\textbf{0.935}} \\
MobileNetV2    & \underline{0.423}  & 0.721  & \underline{0.785}  & 0.863  & 0.775  & 0.909  & 0.514  & \underline{0.733}  & \underline{0.857}  & 0.895 \\
ConvNeXt-S     & 0.371  & 0.557  & 0.611  & 0.496  & 0.621  & 0.806  & 0.314  & \underline{0.733}  & 0.652  & 0.877* \\
ViT-B/16       & 0.278  & 0.361  & 0.367  & 0.841  & 0.474  & 0.731  & 0.235  & 0.541  & 0.442  & 0.855\\ 
Swin-S         & 0.381  & 0.670  & 0.621  & 0.836  & 0.620 & 0.896*  & 0.419  & 0.719  & 0.686  & 0.922* \\ \midrule
Average        & 0.367  & 0.614  & 0.625  & 0.799  & 0.653  & \textbf{0.863}  & 0.412  & 0.682  & 0.681  & \textbf{0.905} \\ \bottomrule
\end{tabular}

\label{tab:cnn}
\end{table*}

\subsection{Analysis of the learned representation} \label{sec:exp_repr_anal}

After training, we choose the best-performing model from the coarse-grained target and extract latent representations from its second-to-last layers. We then apply the UMAP~\cite{umap} dimensionality reduction technique to investigate relationships between the clusters formed in this embedding space, with the primary assumption being that the clusters will follow the class-based separation of the data.

This compressed representation is further compared to the available metadata, specifically to the experimental design of the investigated cheese types. This parameter space is first normalized and expressed with 2-component PCA (visualization available in \emph{Supplementary Material}), resulting in a spatial relationship between classes that can be compared to the UMAP of the clusters. If the models have learned a meaningful representation of the structure, embedding clusters representing similar cheese types should be located close to each other. To enable the visual comparison of UMAP clusters, we assign a color representation to the points in the PCA space.
\section{Results} \label{sec:results}

\subsection{Training}

\begin{figure*}[!t]
\centering
     \begin{subfigure}[b]{0.62\columnwidth}
         \includegraphics[width=\columnwidth]{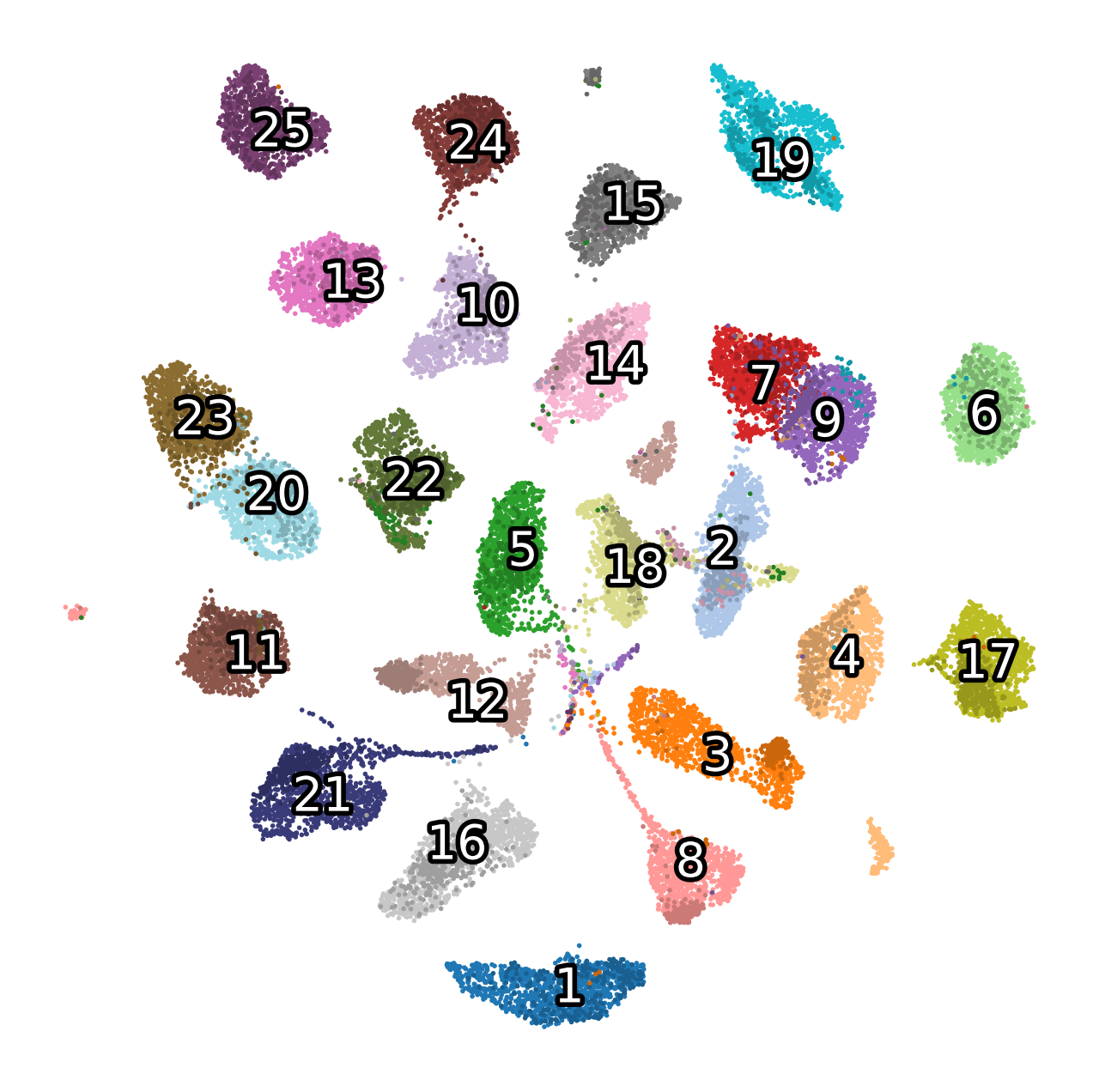}
         \caption{Clusters colored by class.}
         \label{fig:umap_coarse_class}
         \quad
     \end{subfigure}\hspace{1ex}
     \begin{subfigure}[b]{0.62\columnwidth}
         \includegraphics[width=\columnwidth]{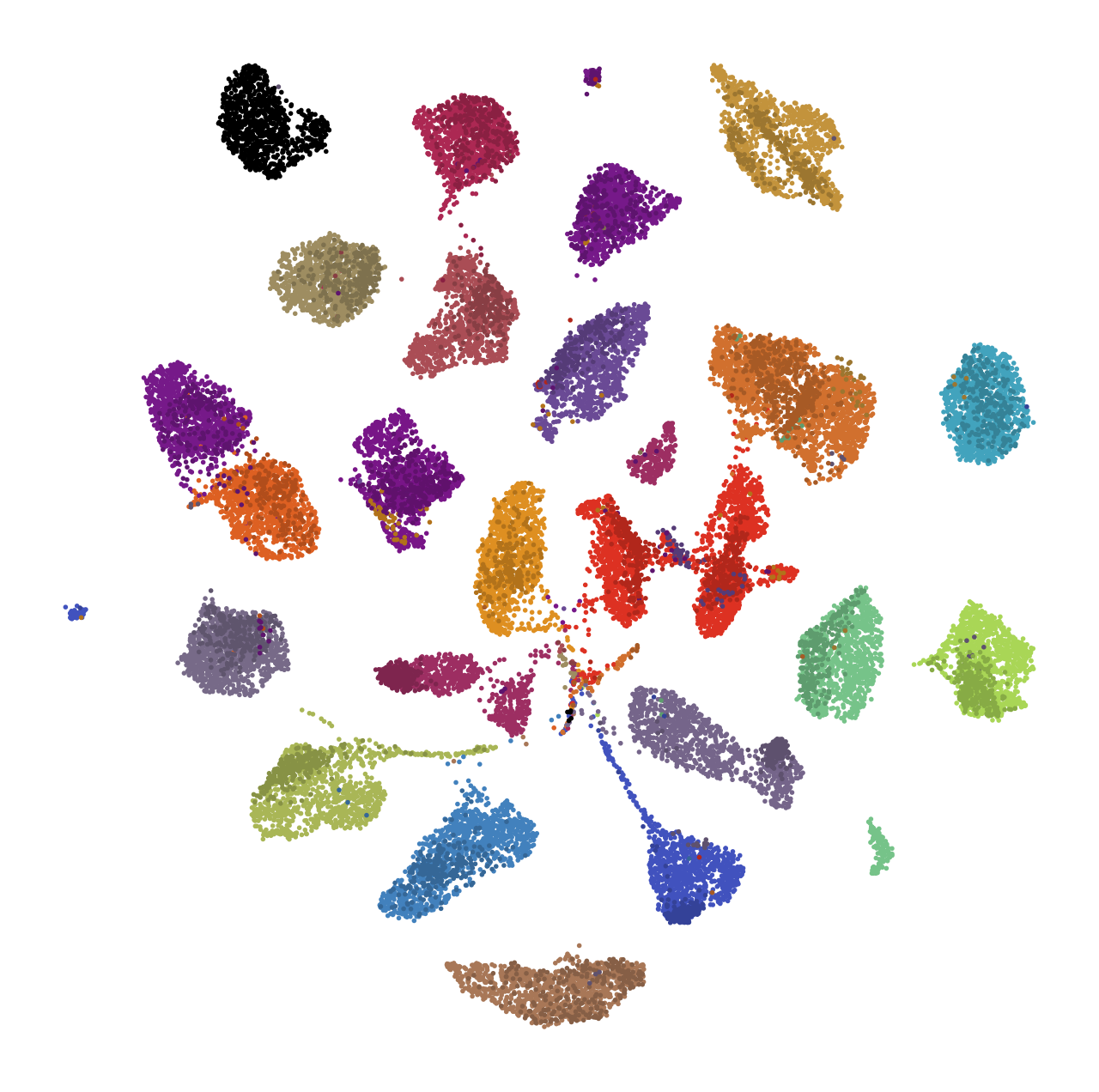}
         \caption{Clusters colored by experimental design PCA color space. Class 25 is colored black.}
         \label{fig:umap_coarse_pca}
     \end{subfigure}\hspace{3ex}
     \begin{subfigure}[b]{0.59\columnwidth}
         \includegraphics[width=\columnwidth]{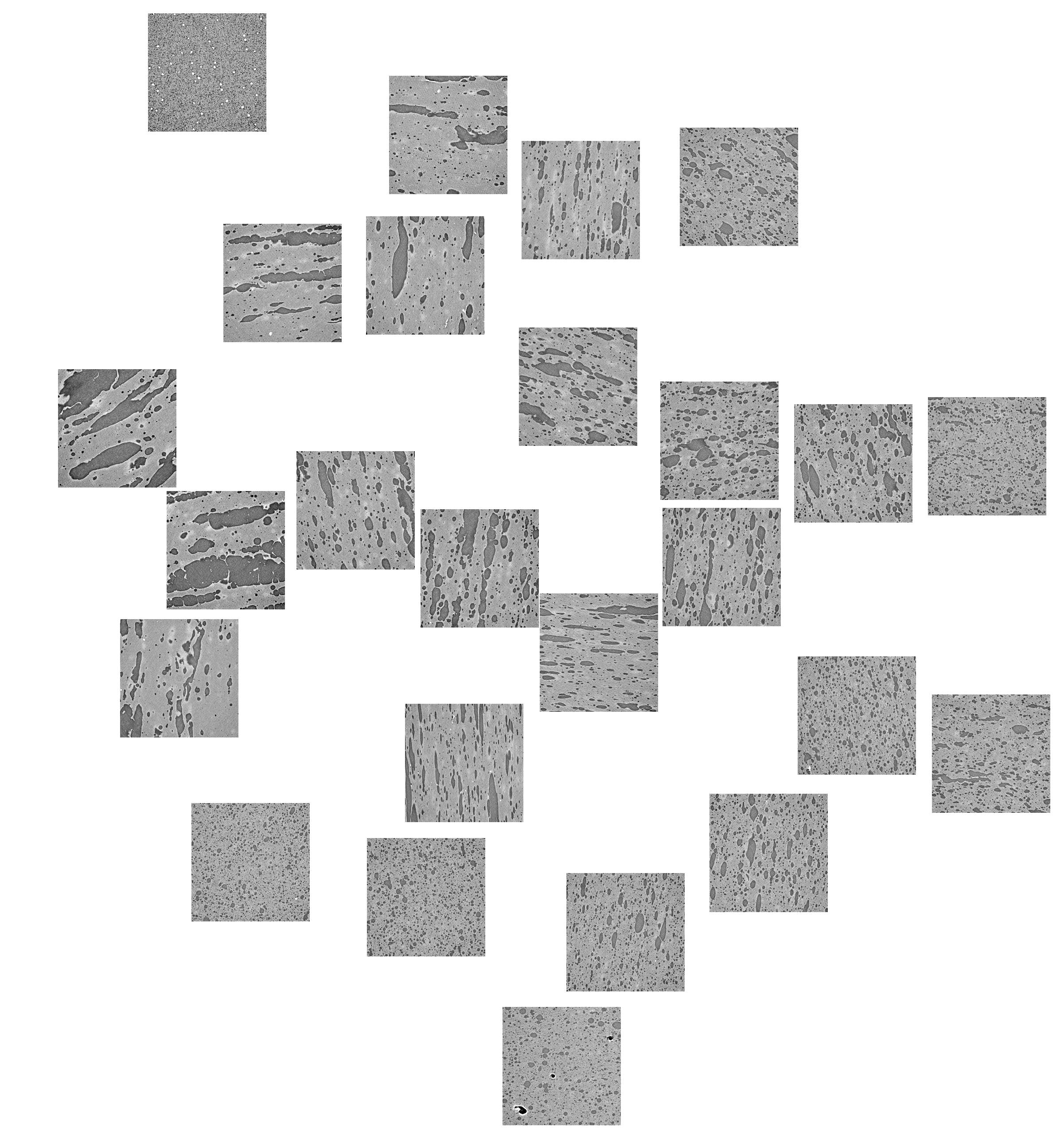}
         \caption{Clusters substituted by example slices from each class.}
         \label{fig:umap_coarse_slice}
     \end{subfigure}
\caption{UMAP generated from second-to-last layer feature representations of the best-performing model in the coarse-grained classification task (ResNet50 trained on the Large dataset). Reduction parameters: \pyth{n\_neighbors=15}, \pyth{min\_dist=0.5}.}
\label{fig:umaps}
\end{figure*}

The majority of the models converged and reached the early stopping criterion, except for 3D Large instances of the ConvNeXt models and the coarse Swin model, which continued to slowly improve until the predefined time limit. Models trained on the Small dataset exhibited strong overfitting to the training data, as expected given the dataset size, though it was partially mitigated through fine-tuning. Notably, the ResNet, MobileNet, and Swin models were the least affected. 

\cref{tab:cnn} shows the complete classification results. Both the average and the per-model metrics demonstrate a significant improvement in accuracy with the 3D data. Interestingly, most 3D models trained using the Coarse-Base setup managed to outperform their counterparts trained on the 2D Large dataset. This suggests that even a smaller 3D dataset is more advantageous than a big 2D dataset for this problem. However, this trend is not visible in the fine-grained classification, indicating that in this task, the image resolution may play a more critical role than the 3D representation.

Across models, ResNet performed the most consistently in all the tasks and achieved the highest accuracy scores for nearly all 3D configurations. MobileNetV2 also demonstrated stable performance, often achieving the highest accuracy among the 2D models. Swin provided a competitive but usually slightly worse performance across all cases, while ConvNeXt and ViT performed the worst, showing significant variability across different configurations.

At a high level, top scores for both granularities (coarse-grained: $0.973$, fine-grained: $0.935$) confirm the validity and feasibility of the classification task. Both scores were only achieved on the Large dataset instance; fine-grained classification was much more challenging in the Base dataset ($0.733$), and coarse-grained was especially problematic in the Small dataset ($0.763$).

From the structural analysis perspective, the high accuracy achieved in the Large dataset suggests that even small volumes, approximately \SI{0.2}{\milli\metre} in size, contain sufficient structural features for accurate classification of cheese types. At the same time, these volumes exhibit enough unique characteristics to identify them as part of a specific sample and its subtle, localized structural variations.

\subsection{Learned representation} \label{sec:res_learned_repr}

UMAP representation of the best-performing coarse-grained model shows distinct clustering of volumes corresponding to each class, with most classes forming tight, well-defined clusters (\cref{fig:umap_coarse_class}). This clustering supports the high accuracy of the model, validating its ability to capture and distinguish between structural variations across classes.

Applying the PCA-based colormap (see \emph{Supplementary Material}) provides additional insights into the UMAP representation (\cref{fig:umap_coarse_pca}). Classes 2, 5, 7, 9 and 18 form a large, central group of clusters with similar colors, suggesting the model has correctly identified their similarity. This group transitions to two additional groups: purple clusters on the top (classes 10, 14, 15, 22) and green/blue clusters on the right (classes 4, 6, 17). The color transition from red to green and red to purple aligns well with the spatial relationships present in the PCA space.

The positioning of some classes (1, 13, 20, and 21) deviates from the alignment with the PCA space, suggesting that the model's representation may be coincidental or at least not very accurate. To further evaluate this deviation, we use a visualization from \cref{fig:umap_coarse_slice}, where the class clusters are substituted with example 2D slices of the volumetric data. This visualization highlights a strong similarity of cheese types located close to each other. Volumes in the center of the map generally display anisotropic structures with large fat domains that grow even further toward the left. Conversely, the right and bottom areas contain volumes with smaller fats and potentially more isotropic structures. With this view, it is clear that the classes deemed problematic in the context of the cheese experimental design are positioned correctly in terms of their actual structural features.
\section{Discussion and Limitations} \label{sec:discussion}

\subsection{Data} \label{sec:disc_data}

The reliability of a dataset depends critically on the quality of its data sources and the errors and noise that may arise at all stages of data generation. In the context of this study, these sources can be summarized into three groups: mozzarella cheese preparation, imaging, and post-processing. 

Mozzarella production, as well as food production in general, cannot be fully controlled, resulting in variations in the final samples. These were highlighted in \cref{sec:res_learned_repr}, where cheese types similar in terms of the experimental design had vastly different structures. While this discrepancy is interesting from a food science perspective, it has a negligible effect on dataset use in deep learning research.
All samples were prepared simultaneously and stored in similar conditions, excluding Cagliata cheese, which is approximately one year older. With age, it developed multiple salt crystals within its matrix, which then created very bright spots on the scans. This effect likely makes it easier to classify this cheese type, but does not compromise the overall dataset.

The power and coherence of synchrotron radiation allowed for generating low-noise and high-contrast scans, and the fast scanning time eliminated the risk of sample movement/deterioration during imaging. Post-processing steps were designed to limit the sources of bias and error, including discarding problematic scans, cropping volumes to avoid artifacts and inconsistencies near the scan edges, and normalizing voxel intensities across samples. 

The orientation of the structure in the samples is also a potential source of bias. If two similar anisotropic structures are imaged with a consistent, different orientation, a model
may learn to classify them based on this orientation alone. The issue is partially mitigated by the random flipping, as well as sample preparation, which involved cutting the samples in varying directions. However, it is not systematically addressed (\eg, through random rotation offsets during reconstruction). To assess the potential impact of this issue, we conduct an additional rotation ablation study, detailed in the \emph{Supplementary Material}. The results show no evidence of orientation bias in the coarse-grained task and a minor negative influence on the fine-grained task.

\subsection{Experiments}
\label{sec:disc_exp}

The high accuracy scores achieved for both classification targets validate the problem's feasibility but may also limit the potential for future improvement. However, the high scores apply only to the largest dataset instance, which is not representative of typical volumetric datasets. Reaching volume counts of over \num{10000} will continue to be problematic for most volumetric imaging tasks, and the size of this dataset instance (\num{37760}) is highly uncommon. 

The classification accuracy drops significantly in the Base and Small instances that reflect more realistic conditions of volumetric datasets. This performance gap is the most prevalent trend in the training result, caused either by the general lack of data or consequent overfitting to training data. We propose that future research could use the biggest instance to establish a base performance of a model, followed by a shift to the smaller data instances that are more challenging and representative of the typical volumetric task. For example, researchers could use the Small dataset with coarse-grained classes or the Base dataset with fine-grained classes.

The difference in performance between 2D and 3D models underscores the importance and impact of volumetric representation for this problem. 
Moreover, the strong performance of ResNet, as well as the discrepancy in MobileNet 2D and 3D results suggest that simpler architectures continue to be the most effective for volumetric data. This further implies that some of the most advanced SotA models may be overly optimized for 2D images, limiting their effectiveness in volumetric tasks and presenting an opportunity to develop models specifically tailored for volumetric deep learning. Notably, the Swin model behavior reinforces this observation --- despite the sensitivity of transformer architectures to dataset size~\cite{vit,lee2021}, it demonstrated surprisingly good performance on the Small and Base splits, making it a promising starting point for future research.

\subsection{Classification as a method of structural analysis}

While classification is a useful tool for high-level data analysis, it may not be the most direct approach for studying fine-grained structures. At the same time, many traditional image analysis methods are targeted specifically for texture and structure. However, some of these methods are very basic~\cite{haralick1973a, granlund1978a}, while others require careful parameter tuning and intensive computation~\cite{Jeppesen2021,stssPieta2024}.

Through this classification task, we aim to explore the broad relationships and tendencies in mozzarella microstructure with implications for the analysis of structured foods. The disordered nature of food structures introduces significant uncertainty in any local measurements. However, our results demonstrate that these local, irregular structures still retain enough distinctiveness to identify their properties both on the coarse and fine-grained levels. Most importantly, these observations underscore the applicability of 3D imaging for detailed and accurate quantification of food structures, motivating its broader application in both research and production.

\subsection{MozzaVID use as a benchmark}

In MozzaVID we decide to focus on classification --- mostly due to practical constraints of the data, as alternative targets would either be impossible or too trivial. This stands in contrast to many other datasets (\cref{tab:datasets,tab:datasets_2D}) that often focus on segmentation or include multiple prediction targets. While this could be seen as a limitation, it is also what lets us effectively perform the proposed data splits (\cref{fig:daatset_instance_creation}) --- a defining property of MozzaVID. Through these splits, we are able to generate an unprecedented amount of volume samples, and to provide a complex fine-grained task with more classes than any existing volumetric dataset (\cref{tab:datasets}). At the same time, classification remains the simplest, most fundamental task, enabling fast experimentation and more direct investigation of model performance.

An important observation emerges from comparing research trends in 2D and volumetric deep learning. Recent advancements in 2D model development suggest the superiority of transformer-based architectures~\cite{xie2021b,cheng2022a,vit,swin,maur2023a}, and there have been promising efforts to apply transformers for specific volumetric data~\cite{hatamizadeh2022a,Chen2024,chen2021vitvnet,xie2021a}. However, volumetric deep learning challenges continue to be dominated by more established convolution-based models~\cite{heller2021a,bilic2023a,setio2017a}. This same trend is also evident in our experiments (\cref{sec:results,sec:disc_exp}), suggesting that future benchmarking results achieved on MozzaVID can serve as an indication of model performance across the field. 
\section{Conclusion} \label{sec:conclusion}

The MozzaVID dataset, with its substantial size and versatility, offers a flexible framework for volumetric model evaluation. By bridging the gap between established 2D benchmarks and existing volumetric datasets, MozzaVID facilitates the development of models tailored specifically for volumetric data. Models trained with MozzaVID provide valuable insights into the properties and patterns of mozzarella microstructure, setting the stage for more advanced structural analysis, both in mozzarella and other structured foods. The dataset can be explored through: \url{https://papieta.github.io/MozzaVID/}

\vspace{6pt}
\noindent \textbf{Acknowledgments}: This work was supported by Innovation Fund Denmark, project 0223-00041B (ExCheQuER). We acknowledge the MAX IV Laboratory for beamtime on the DanMAX beamline under proposal 20240276. 
\clearpage
{
    \small
    \bibliographystyle{ieeenat_fullname}
    \bibliography{main}
}

\clearpage
\maketitlesupplementary

\section{Train/validation/test splitting strategy}

The splitting strategy for the mozzaVID dataset is tailored for the specific requirements of both classification targets, resulting in two separate approaches.

The coarse target is designed to explore high-level structural differences induced by the cheese recipe parameters. However, it is assumed that volumes from the same scan, or the same sample, are physically dependent due to the proximity of the imaged structures. Thus, the splitting was done on a sample level to ensure independence between volumes in each split. Since there are only 6 samples per scan, the train/val/test split was done in ratios of approximately 67\%, 16\%, and 16\%, respectively.

In contrast, models trained on the 150 classes of the fine target are expected to exploit and evaluate the degree of the aforementioned proximity-based structural dependence. Because of that, the splits for fine target were simply performed on individual volume level, with a standard ratio of 70\%, 20\%, and 10\% for train, validation, and test, respectively.

\section{Data visualization}

\subsection{Scans}

\begin{figure*}[!t]
\centering
\includegraphics[width=\textwidth]{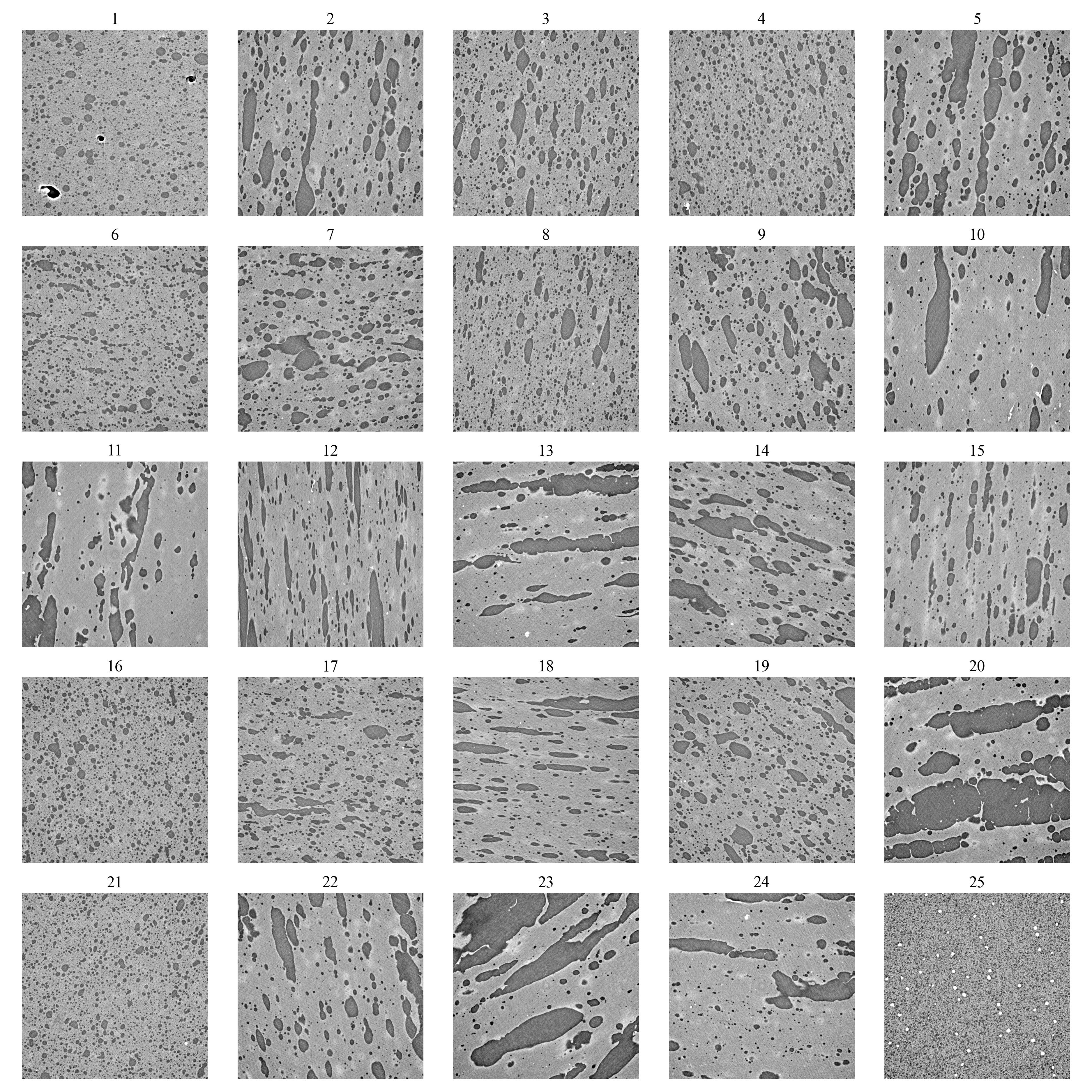}
\caption{
Overview of slices from each cheese type, forming the 25 coarse-grained classes.}
\label{fig:cheese_slices}
\end{figure*}

In \cref{fig:scaning_and_example_scans}, we introduce a set of example scan slices that illustrate the structural variability across the cheese samples. Subsequently, in \cref{fig:umap_coarse_slice}, we use slices from all scans to explore the representation learned by one of the models. To provide a clean and comprehensive overview of the variability, the same slices are arranged in an ordered grid in \cref{fig:cheese_slices}.

\begin{figure*}[t]
\centering
\includegraphics[width=\textwidth]{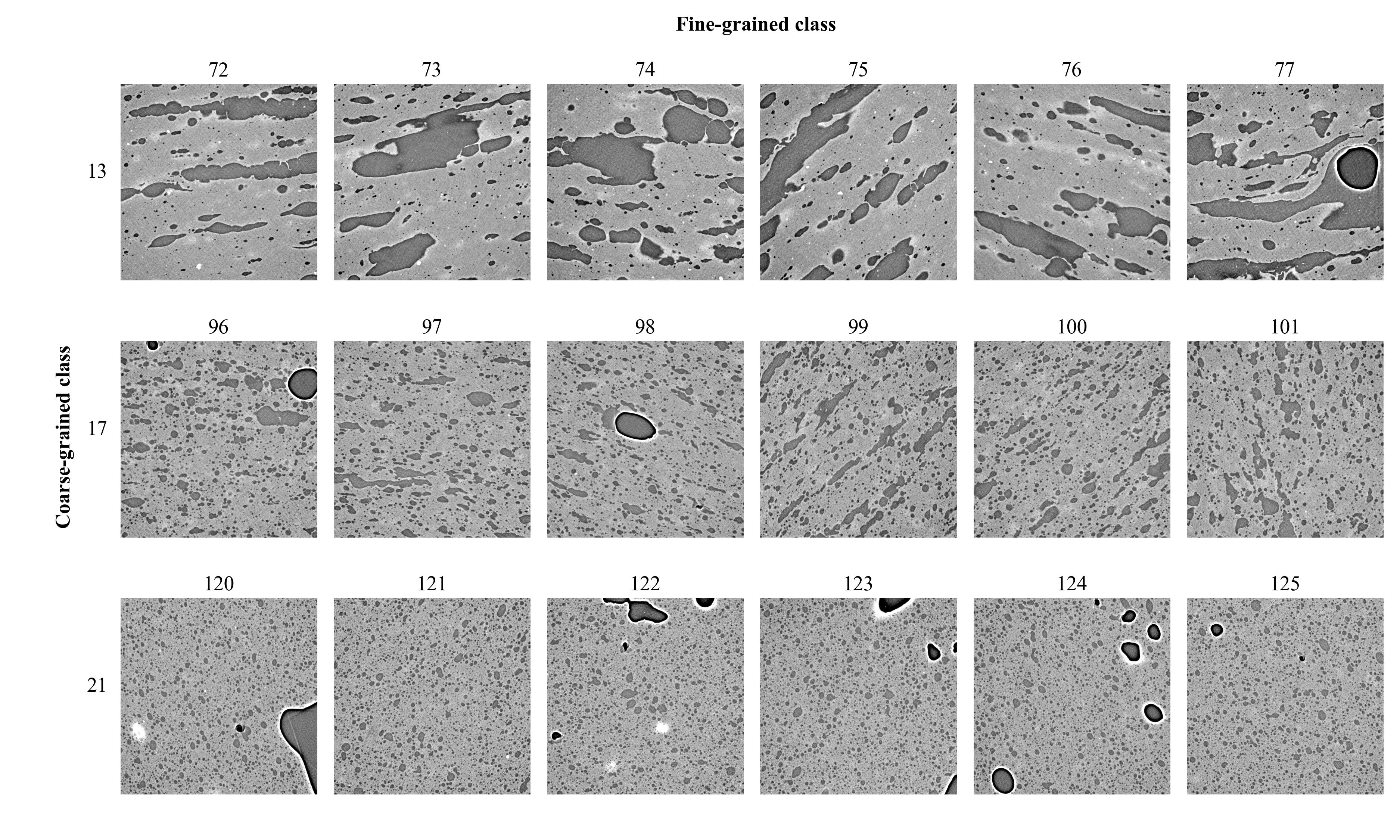}
\caption{
Example slices from the fine-grained classes. Each row represents a set of six samples from one cheese type (coarse-grained class), forming six consecutive fine-grained classes.}
\label{fig:sample_slices}
\end{figure*}

To further supplement the overview of the scanned samples, \cref{fig:sample_slices} showcases example slices from different fine-grained classes. These are organized into sets of six samples originating from the same cheese type, emphasizing their structural similarity and consequent increase in the complexity of the problem. However, it is important to note that a single 2D slice may not fully capture the sample's structural characteristics, which are likely to be more effectively discerned through a full 3D representation.

\subsection{Metadata}

In \cref{sec:data_acq_preproc}, we provide an overview of the metadata, including experimental design parameters, which are later visualized in a reduced form in \cref{fig:exp_design_pca}. 

\begin{figure*}[t]
\centering
\includegraphics[width=\textwidth]{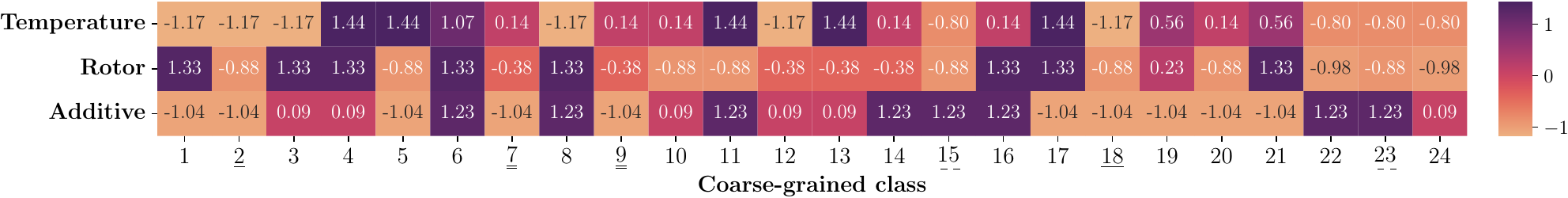}
\caption{
Overview of the variation in the normalized experimental design parameters of the first 24 cheese types (coarse-grained classes). Data for class 25 (Cagliata) is not available. Underlines highlight three pairs of cheese produced with the same set of parameters.}
\label{fig:exp_design_heatmap}
\end{figure*}

The following PCA reduction was applied to three parameters: rotor speed, temperature, and additive type. The additive type is a categorical variable with three categories (None, CaCL2, and Citric Acid), to which a unique number is assigned. Each parameter is then normalized to maintain the confidentiality of the exact recipe. The resulting values are presented in \cref{fig:exp_design_heatmap}, normalized to a zero mean and standard deviation of one.

\section{Experiments}

\subsection{Top-k accuracy}

In~\cref{tab:cnn}, top-1 accuracy is provided as a primary evaluation of model performance. To provide further context, top-k accuracy scores are also listed in~\cref{tab:model_acc_top_k} (top-2 for coarse granularity and top-5 for fine granularity). 

For both coarse and fine models, there is a significant increase in accuracy compared to~\cref{tab:cnn}, with Large 3D models approaching near-perfect performance. This indicates that the models can easily choose a set of the most probable classes, but may struggle to discriminate between a few most similar neighbors.

Importantly, the examination of confusion matrices for both granularities shows no systematic trends or recurring pairs of misclassified classes, suggesting that the space of investigated structures is sufficiently broad and relatively uniformly distributed.

\begin{table*}[t!]
\caption{Top-k accuracy of the trained models. Top-2 accuracy is reported for the coarse granularity and top-5 for the fine granularity, based on the total class count in both targets.}
\centering
\footnotesize
\begin{tabular}{@{}l@{\hskip 20pt}cc|cc|cc@{\hskip 30pt}cc|cc@{}}
\toprule
Granularity   & \multicolumn{6}{c@{\hskip 35pt}}{Coarse, top-2}  & \multicolumn{4}{c}{Fine, top-5} \\ \midrule
Split         & \multicolumn{2}{c}{Small}  & \multicolumn{2}{c}{Base} & \multicolumn{2}{c@{\hskip 30pt}}{Large} & \multicolumn{2}{c}{Base} & \multicolumn{2}{c}{Large} \\ \midrule
Dimensionality & \textbf{2D}     & \textbf{3D}     & \textbf{2D}     & 3D     & \textbf{2D}     & 3D     & \textbf{2D}     & 3D     & \textbf{2D}     & 3D   \\ \midrule
ResNet50       & 0.567 & 0.876 & 0.883 & 0.991 & 0.832 & 0.997  & 0.853 & 0.991  & 0.983 & 0.999 \\
MobileNetV2    & 0.660 & 0.901 & 0.888 & 0.951 & 0.910 & 0.972  & 0.783 & 0.996  & 0.985 & 0.989 \\
ConvNeXt-S     & 0.516 & 0.714 & 0.787 & 0.735 & 0.787 & 0.908  & 0.867 & 0.946  & 0.914 & 0.996 \\
ViT-B/16       & 0.423 & 0.516 & 0.602 & 0.915 & 0.594 & 0.866  & 0.566 & 0.964  & 0.888 & 0.997 \\
Swin-S         & 0.557 & 0.814 & 0.800 & 0.972 & 0.787 & 0.958  & 0.889 & 0.993  & 0.988 & 0.999 \\ \midrule
Average        & 0.545  & 0.764  & 0.792  & 0.920  & 0.782  & 0.940  & 0.792  & 0.978  & 0.952  & 0.996 \\ \bottomrule
\end{tabular}
\normalsize
\label{tab:model_acc_top_k}
\end{table*}

\subsection{Learning rate fine tuning}

In \cref{sec:experiments}, we outline the experimental design, including the investigated models and ablation studies. The training setup across all models is standardized to ensure a fair comparison of the models. However, certain hyperparameters should be fine-tuned to provide the most representative result. In this study, we focus on fine-tuning the learning rate, as it has the most significant impact on the consistency of model performance. Given the training and convergence constraints, fine-tuning was performed only for 2D models and 3D-fine models, while the remaining models used a default learning rate of \num{e-4}.

Fine-tuning was conducted using the \emph{Weights \& Biases} parameter sweep over a log-uniform distribution in the range $lr \in <\num{e-2}, \num{e-6}>$, using the Bayesian optimization. Each model was tested with 30 hyperparameter variants, provided this could be achieved in 1 day of training. Otherwise, the number of tested variants was lowered to 15. The best configuration was selected based on the highest validation accuracy. The final learning rates for all models are listed in \cref{tab:model_lr}.

\begin{table*}[t!]
\caption{Learning rate of the trained models on all investigated setups after fine-tuning. Base and Large 3D models were assigned a default learning rate due to the limited resources and slow convergence.}
\centering
\footnotesize
\begin{tabular}{@{}l@{\hskip 20pt}cc|cc|cc@{\hskip 30pt}cc|cc@{}}
\toprule
Granularity   & \multicolumn{6}{c@{\hskip 35pt}}{Coarse}  & \multicolumn{4}{c}{Fine} \\ \midrule
Split         & \multicolumn{2}{c}{Small}  & \multicolumn{2}{c}{Base} & \multicolumn{2}{c@{\hskip 30pt}}{Large} & \multicolumn{2}{c}{Base} & \multicolumn{2}{c}{Large} \\ \midrule
Dimensionality & \textbf{2D}     & \textbf{3D}     & \textbf{2D}     & 3D     & \textbf{2D}     & 3D     & \textbf{2D}     & 3D     & \textbf{2D}     & 3D   \\ \midrule
ResNet50       & \num{9.9 e-4} & \num{7.3 e-5} & \num{1.1 e-3} & \num{e-4} & \num{2.8 e-3} & \num{e-4} & \num{9.0 e-4} & \num{e-4}  & \num{3.3 e-4} & \emph{\num{e-4}} \\
MobileNetV2    & \num{2.1 e-3} & \num{5.8 e-4} & \num{1.6 e-3} & \num{e-4} & \num{3.3 e-3} & \num{e-4}  & \num{1.9 e-3} & \num{e-4}  & \num{6.0 e-4} & \emph{\num{e-4}} \\
ConvNeXt-S     & \num{3.4 e-3} & \num{1.7 e-3} & \num{2.2 e-3} & \num{e-4} & \num{1.6 e-3} & \num{e-4}  & \num{1.4 e-3} & \num{e-4}  & \num{1.0 e-3} & \emph{\num{e-4}} \\
ViT-B/16       & \num{1.8 e-5} & \num{6.0 e-5} & \num{4.1 e-5} & \num{e-4} & \num{1.6 e-5} & \num{e-4}  & \num{3.4 e-6} & \num{e-4}  & \num{5.0 e-5} & \emph{\num{e-4}} \\
Swin-S         & \num{1.8 e-4} & \num{5.4 e-5} & \num{1.2 e-4} & \num{e-4} & \num{2.9 e-5} & \num{e-4}  & \num{2.4 e-6} & \num{e-4}  & \num{8.8 e-5} & \emph{\num{e-4}} \\ \bottomrule
\end{tabular}
\normalsize
\label{tab:model_lr}
\end{table*}

Although not all models underwent fine-tuning, those that did were also the ones most likely to benefit from it. The performance of the tested models tended to be more volatile and sensitive when trained on the smaller dataset variants. In the case of 2D-Small models, small learning rate changes often resulted in a $20$ to $30$ percentage point difference in accuracy, while for 2D-Large models, this difference was limited to approximately $1$ to $5$ percentage points. This behaviour suggests that the performance of 3D models, even without fine-tuning, remains representative and close to optimal. Furthermore, the strong performance of non-fine-tuned 3D models only strengthens the reported impact of the 3D representation and all conclusions drawn from it.

\subsection{PCA visualization}

\begin{figure}[!t]
\centering
\includegraphics[width=\columnwidth]{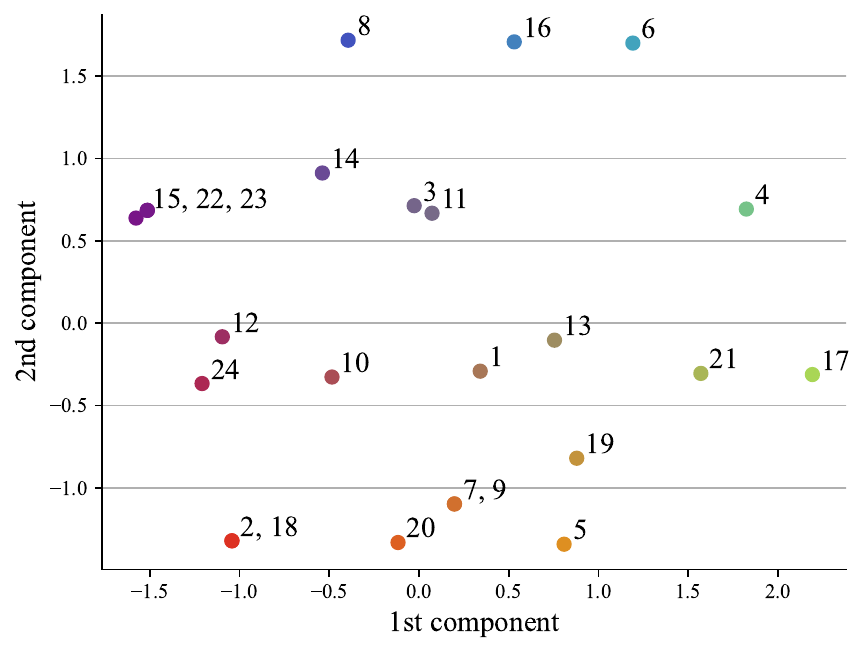}
\caption{
PCA of the experimental design parameters used to define the mozzarella cheese variations. Class 25 (Cagliata cheese) is omitted -- it was prepared separately, and its exact recipe is not available. Coloring of the points serves as a basis for visual analysis of structural similarities learned by the models (\cref{fig:umap_coarse_pca}).}
\label{fig:exp_design_pca}
\end{figure}

In \cref{sec:exp_repr_anal}, we introduce the outline of the learned representation analysis experiment. There, a PCA-based compression of the sample experimental design parameter space is described, together with an introduction of a colormap that is later used for the interpretation of the UMAP representation in \cref{fig:umap_coarse_pca}. The visualization of the PCA space, together with the colormap overlay, can be found in \cref{fig:exp_design_pca}.

\subsection{Rotation ablation study}

In \cref{sec:disc_data}, we discuss structure orientation as a potential source of bias, noting that while sample preparation and data augmentations largely mitigate this issue, it is not systematically addressed in the raw data.  To evaluate the impact of volume orientation on the accuracy of the investigated models, we conducted an additional ablation study using a modified set of transforms.

This study focused on the ResNet50 model, given its consistent performance, and utilized the Base dataset as a central, representative example. Both 2D and 3D models were trained for coarse-grained and fine-grained tasks. Apart from the data augmentation, the training setup was exactly the same as in the original experiments.

\newpage

The modified set of transforms included the following elements:
\begin{enumerate}
    \item Normalization.
    \item Random \ang{90} rotation ($p=0.5$).
    \item Random flipping in X and Y axis ($p=0.5$).
    \item Random rotation in a \qtyrange{-30}{30}{\degree} range ($p=0.5$).
\end{enumerate}
Elements 1 and 3 were retained from the original pipeline. The combined set of transforms covers most of the possible structure orientations while minimizing information loss at the most extreme angles.

\begin{table}[t!]
\caption{Results of rotation ablation study using the ResNet50 model and the Base dataset instance. Reference results copied from \cref{tab:cnn}.}
\centering
\begin{tabularx}{\columnwidth}{@{}l@{\hskip 30pt}cc|cc}
\toprule
Granularity   & \multicolumn{2}{c}{Coarse}  & \multicolumn{2}{c}{Fine} \\ \midrule
Dimensionality & 2D     & 3D     & 2D     & 3D      \\ \midrule
Reference    & 0.741  & 0.957  & 0.563  & 0.683 \\
With rotations       & 0.747  & 0.945 & 0.498  & 0.643 \\ \bottomrule
\end{tabularx}
\label{tab:rot_study}
\end{table}

The results of the study (\cref{tab:rot_study}) suggest that volume orientation does not exhibit a clear or consistent impact on training accuracy. The coarse-grained models seem to perform similarly with the new setup. The impact on fine-grained models is more negative compared to coarse-grained models, which may be due to fine-grained models relying more heavily on structure orientation, as each sample (fine-grained class) is cut in a single direction, with variation introduced only through data augmentation. However, this effect is not significant enough to undermine the validity of the dataset or the presented results.

\begin{figure*}[!t]
\centering
     \begin{subfigure}[b]{0.45\textwidth}
         \includegraphics[width=\columnwidth]{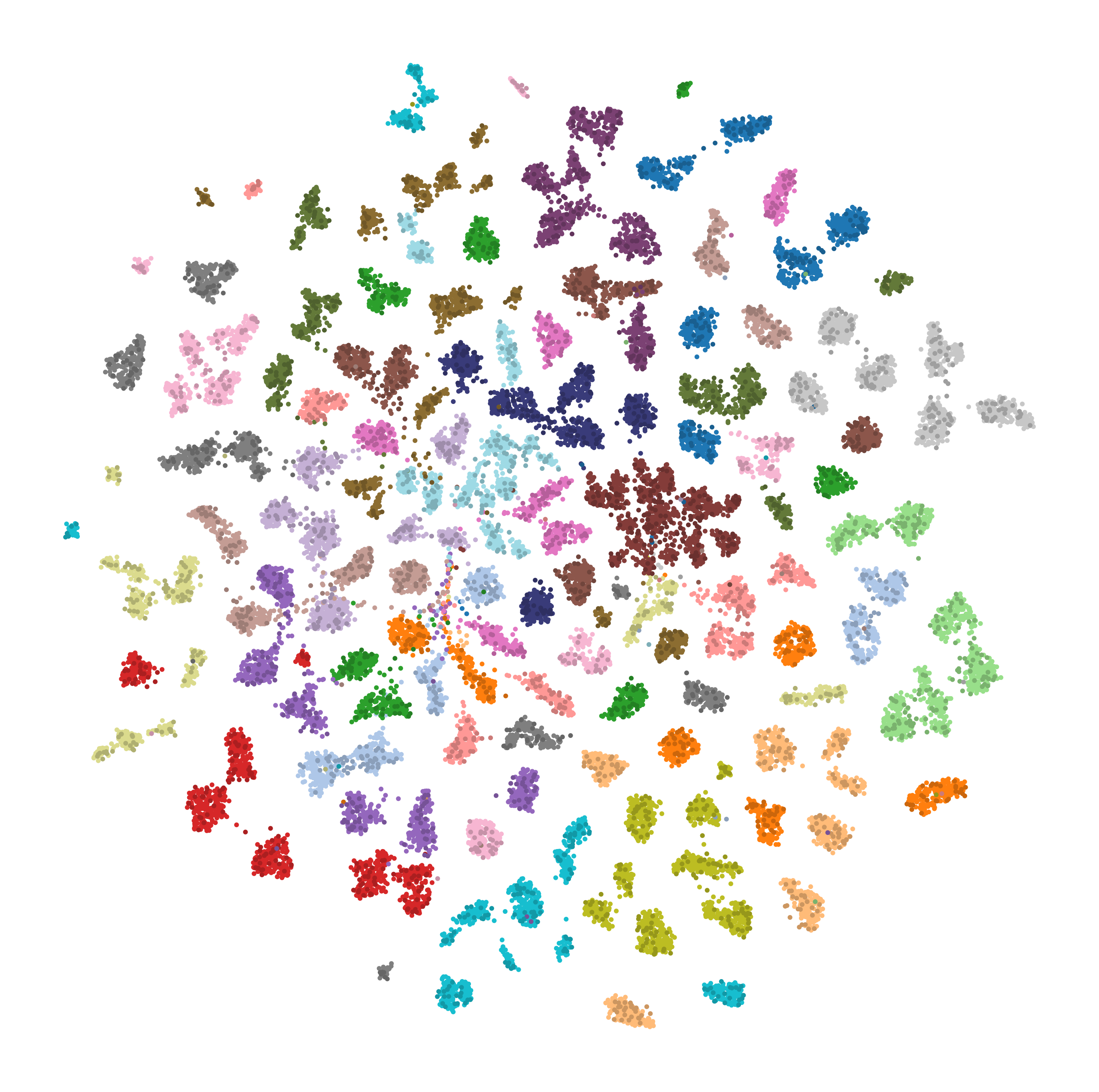}
         \caption{Clusters colored by coarse-grained class.}
         \label{fig:umap_fine_class}
         \quad
     \end{subfigure}
     \begin{subfigure}[b]{0.45\textwidth}
         \includegraphics[width=\columnwidth]{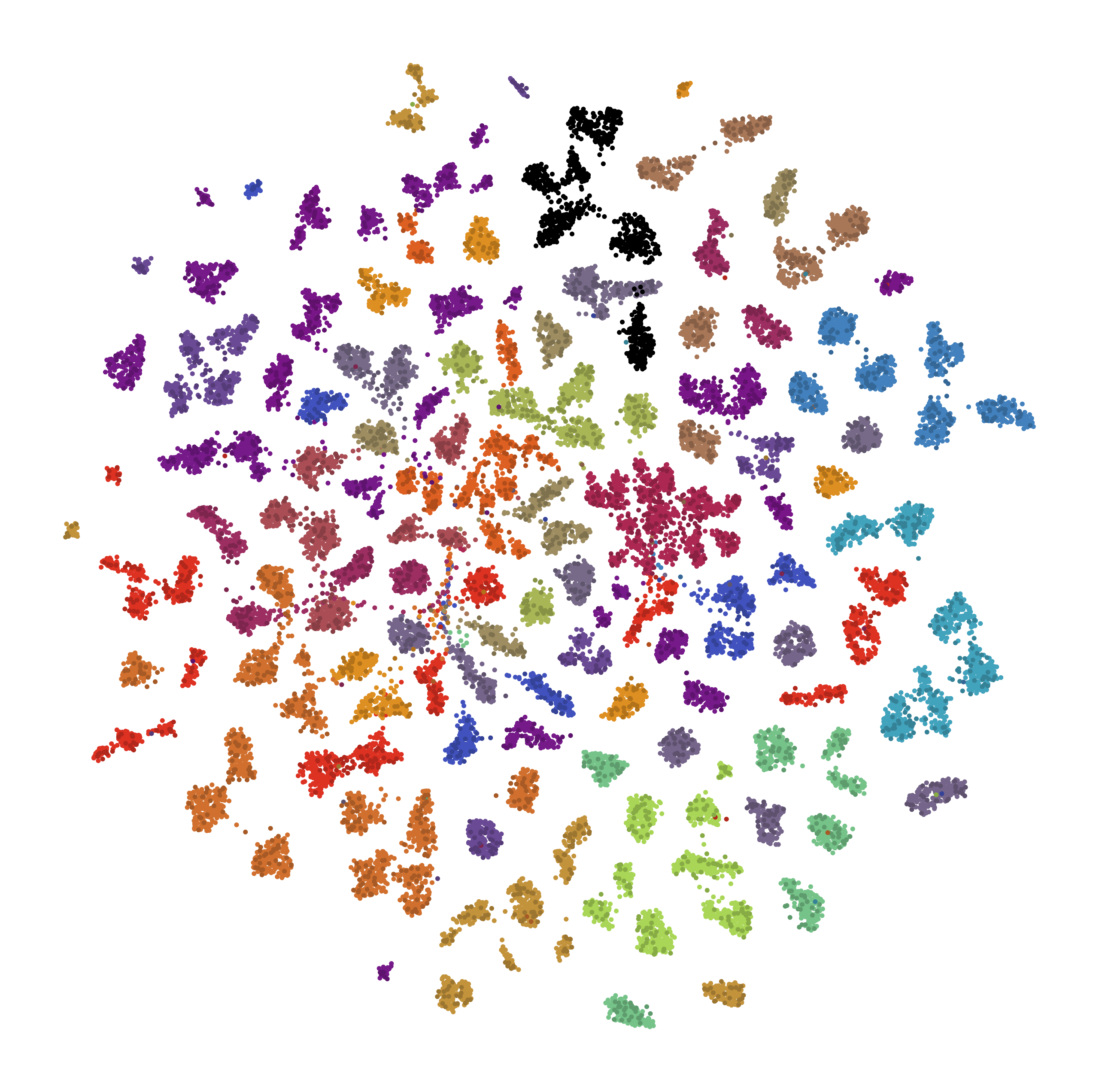}
         \caption{Clusters colored by experimental design PCA color space. Class 25 is colored black.}
         \label{fig:umap_fine_pca}
     \end{subfigure}
\caption{UMAP generated from second-to-last layer feature representations of the best-performing model in the fine-grained classification task (ResNet50 trained on the Large dataset). Reduction parameters: \pyth{n\_neighbors=30}, \pyth{min\_dist=0.8}.}
\label{fig:umaps_fine}
\end{figure*}

\subsection{Fine-grained model UMAP}

A similar experiment to the one described in \cref{sec:exp_repr_anal} and visualized in \cref{fig:umaps} was conducted using the best-performing fine-grained model. The resulting UMAPs are shown in \cref{fig:umaps_fine}, though the slice-based visualization is omitted here for readability. The colormaps remain consistent with those used in the coarse-grained analysis.

Despite a different target, in most cases the network still groups samples from the same coarse-grained class in close proximity (\cref{fig:umap_fine_class}). This behavior indicates that structural similarities within these samples significantly influence the embedding space constructed by the model. As shown in \cref{fig:umap_fine_pca}, the alignment with PCA space appears similar or even stronger compared to the coarse-grained model. The UMAP reveals four distinct zones with purple samples in the top-left, orange/red in the bottom-left, green in the bottom-right, and blue on the right.  While some exceptions are present, these can often be explained by structural properties that deviate from the PCA parameters, as discussed in \cref{sec:res_learned_repr}. This layout suggests that the fine-grained model captures a more nuanced representation of the structural variability, which may result from the need to detect subtler differences between samples and the absence of regularizing constraints imposed by the 25 coarse-grained classes.

\end{document}